\documentclass[runningheads]{llncs}
\usepackage[T1]{fontenc}
\usepackage{cite}
\usepackage{amsmath,amssymb,amsfonts}
\usepackage{algorithmic}
\usepackage{graphicx}
\usepackage{textcomp}
\usepackage{xcolor}
\usepackage{multirow}
\usepackage{url}
\begin{document}

\title{A New Error Temporal Difference Algorithm for Deep Reinforcement Learning in Microgrid Optimization\thanks{This work was supported in part by the Royal Society under grants IEC\textbackslash NSFC\textbackslash 201107.}}
%
%
\author{Fulong Yao* \and
Wanqing Zhao \and
Matthew Forshaw}
\authorrunning{Fulong Yao et al.}
%
\institute{{School of Computing, Newcastle University, Newcastle upon Tyne, NE4 5TG, UK
\email{*Corresponding author: f.yao3@ncl.ac.uk}}}
\maketitle              
\begin{abstract}
Predictive control approaches based on deep reinforcement learning (DRL) have gained significant attention in microgrid energy optimization. However, existing research often overlooks the issue of uncertainty stemming from imperfect prediction models, which can lead to suboptimal control strategies. This paper presents a new error temporal difference (ETD) algorithm for DRL to address the uncertainty in predictions, aiming to improve the performance of microgrid operations. First, a microgrid system integrated with renewable energy sources (RES) and energy storage systems (ESS), along with its Markov decision process (MDP), is modelled. Second, a predictive control approach based on a deep Q network (DQN) is presented, in which a weighted average algorithm and a new ETD algorithm are designed to quantify and address the prediction uncertainty, respectively. Finally, simulations on a real-world US dataset suggest that the developed ETD effectively improves the performance of DRL in optimizing microgrid operations. 

\end{abstract}
\section{Introduction}\label{sec1}
\vspace{-5pt} 
The global energy landscape is undergoing a significant transformation driven by increasing energy demands, the need for carbon reduction and the integration of renewable energy sources (RES) \cite{1-1}. This requires developing efficient optimization solutions for the operation of energy systems to ensure their sustainability and cost-effectiveness. Microgrids, which integrate RES (e.g., photovoltaic (PV)) and energy storage system (ESS) with traditional power systems, offer a promising solution to enhancing energy efficiency and resilience. However, the inherent variability and uncertain RES generations pose serious challenges to the optimization of microgrids \cite{1-3}. 

In the past few years, various methods have been developed to optimise energy management in microgrids. For example, Franke et al. \cite{1-4} developed a stochastic optimization model based on mixed integer linear programming (MILP) for a hybrid residential microgrid system in Germany, to enhance both heat and electricity utilization. Torkan et al. \cite{1-5} developed an innovative optimization approach based on genetic algorithm (GA) and stochastic programming for a district microgrid. Considering reactive loads, demand response programs and uncertainties from RES, this approach effectively reduced operational costs while curbing pollution. Raghavan et al. \cite{1-6} introduced a day-ahead scheduling strategy for ESS based on GA and particle swarm optimization (PSO), aiming to minimize operational costs under forecasted microgrid variables (e.g., load, energy generation and electricity price). These methods have demonstrated impressive performance in optimizing the energy flows within a microgrid. However, such methods have demonstrated sensitivity to parameter tuning and face challenges in converging to optimal solutions for microgrid control \cite{1-7}.

Recently, reinforcement learning (RL), particularly deep reinforcement learning (DRL), has demonstrated distinct advantages in optimization problems due to its powerful learning and generalization capabilities. The strong adaptability and exploration features of RL and DRL allow them to better work in the dynamic and uncertain microgrid environments. Huang et al. \cite{1-8} proposed a DQN-based microgrid formation method to search for the optimal control policy in a model-free fashion, significantly improving the resilience of distributed microgrids. Wu et al. \cite{1-9} designed a double DQN (DDQN) method for energy trading and demands scheduling among multiple microgrid systems. When using DRL to optimize microgrid operations, near-future predictions such as electricity price to better inform the design of control strategies. This has resulted in an a multitude of DRL- based predictive control approaches. For instance, Cao et al. \cite{1-10} presented a predictive control method to minimise battery degradation and operational costs. A hybrid convolutional neural network-LSTM (H-CNN-LSTM) was introduced to forecast the electricity prices for the next 24 hours, while a Noisy Net-DDQN (NN-DDQN) optimization was proposed to control battery behaviours based on these predicted prices. Harrold et al. \cite{1-11} proposed a DRL-based predictive control approach aimed at reducing operational costs and optimizing the utilization of RES generations. A regression ANN was used to forecast the near-future values of prices, demands and RES generations, while an actor-critic method was employed to enhance the microgrid performance. 

Although existing works have suggested the effectiveness of DRL-based predictive controls in microgrid optimization, there remains a critical gap that requires further research. As indicated, the predicted values are used as inputs of the DRL to infer the optimal control policy. However, these predictions are naturally accompanied with uncertainties. Unfortunately, existing research in predictive control often neglects such uncertainties, resulting in suboptimal control strategies. In our recent works, two prediction models, a convolutional neural network- long short term memory (CNN-LSTM)\cite{1-12} and a self-organizing interval type-2 fuzzy neural network with multiple outputs (SOIT2FNN-MO) \cite{1-13}, have been designed for time series prediction problem. Building on these efforts, this paper proposes a weighted average algorithm and a new error temporal difference (ETD) algorithm to quantify and address the prediction uncertainty, aiming to improve the optimization performance in DRL (e.g., DQN)-based predictive controls. The main contributions can be listed as follows: 

1) A microgrid system integrated with renewable energy sources (RES) and energy storage systems (ESS) is modelled, along with its Markov decision process (MDP); 

2) A predictive control approach based on a deep Q network (DQN) is presented, incorporating a weighted average algorithm to quantify prediction uncertainty and a novel ETD algorithm to address it;

3) The effectiveness of the proposed approach is validated using a US microgrid with the aim of reducing operational costs and carbon emissions. 

The remaining sections of this paper are organized as follows. Section \ref{sec2} introduces and models a microgrid integrated with RES and ESS. Section \ref{sec3} details the proposed approach. Section \ref{sec4} presents some simulations to demonstrate the effectiveness of our algorithm, while Section \ref{sec5} concludes the paper and outlines areas for future work.

\section{Microgrid Description and Its Modelling}\label{sec2}
\subsection{Microgrid Modelling}\label{sec2.1}
This paper focuses on DRL for battery controls in a broad range of grid-connected microgrid systems incorporating RES and ESS, as illustrated in Fig. \ref{fig1}. Here, the energy supply is either from the connected grid or local RES. A battery device is deployed to temporarily store the excess RES generations (after meeting the demand) for later use. Besides, the battery can also act as a buffer by importing electricity from the grid when the price and carbon intensity are low, and then discharging to meet the demand when the price and carbon intensity are high. Furthermore, we assume that the microgrid has not joined any existing prosumer schemes, i.e., surplus RES production is not sent back to the grid. In other words, if RES production exceeds the demand and the battery is fully charged, the surplus will be curtailed and discarded. In such a microgrid, the electricity balance between consumption and supply must always be met:
\begin{figure}[htbp]
	\vspace{-22pt} 
	\centerline{\includegraphics[width=7cm]{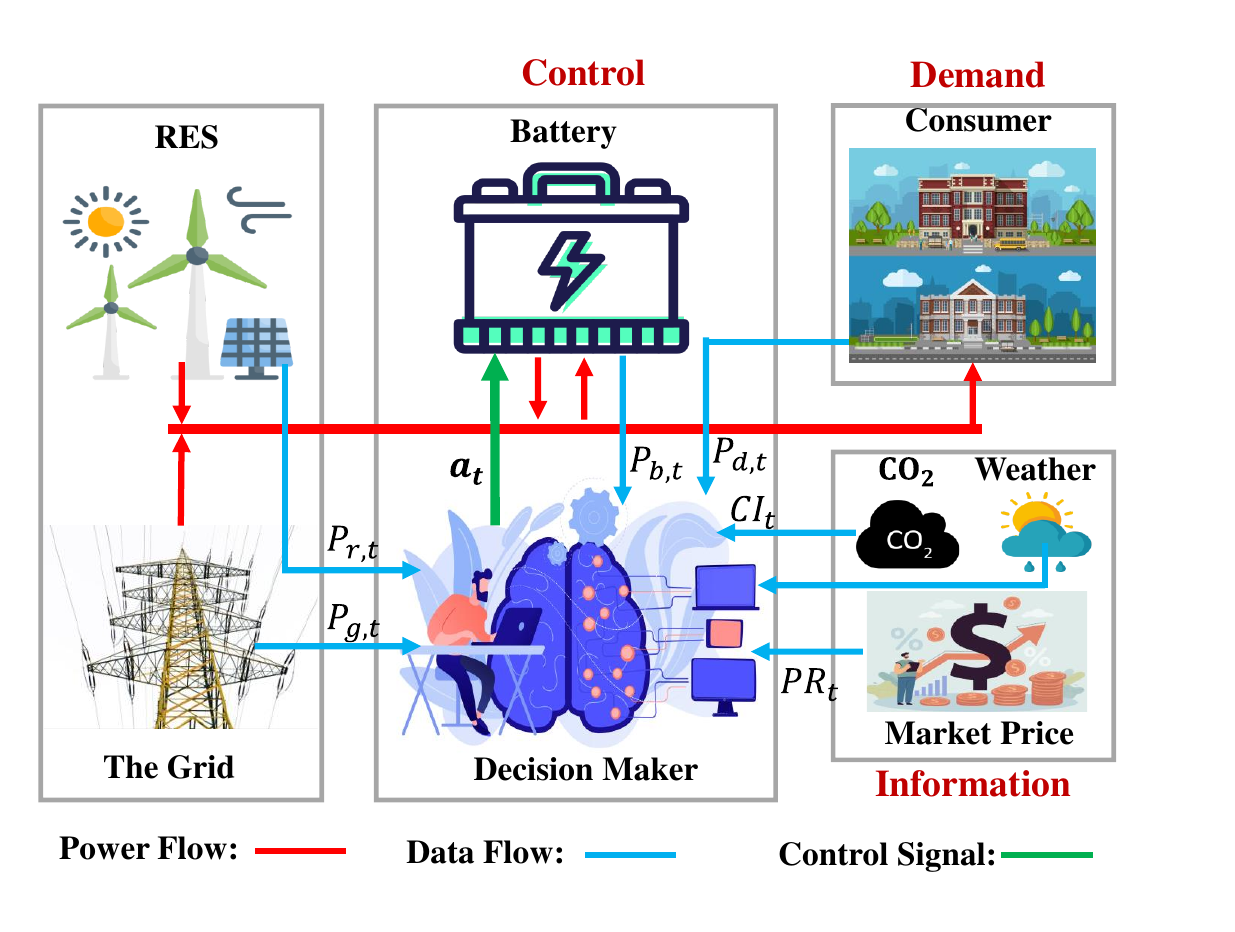}}
	\vspace{-12pt} 
	\caption{Schematic of the microgrid}
	\label{fig1}
	\vspace{-20pt} 
\end{figure}
\begin{equation} \label{eq:1}
	\vspace{-3pt} 
	P_{d,t}  + P_{b,t} - P_{r,t} - P_{g,t}+ P_{c,t} = 0
	\vspace{-1pt} 
\end{equation}
where $ P_{d,t}\geq0 $ (kW) is the total electricity demand at time instant $ t $, $ P_{b,t} $ (kW) is the power discharged ($ P_{b,t}<0 $) or charged ($ P_{b,t}>0 $) by the battery, $ P_{r,t}\geq0 $ (kW) is the RES production, $ P_{g,t}\geq0 $ (kW)  is the imported power from the grid and $ P_{c,t}\geq0 $ (kW) is the curtailed power. To simplify the problem statement, we define the unmet power as: $ P_{u,t}=P_{d,t}-P_{r,t} $. Here, $ P_{u,t}<0 $ (kW) indicates that the RES production can meet the microgrid's demand, while $ P_{u,t}>0 $ indicates that additional power must be imported from the grid to satisfy the demand. Then, the following holds:
\begin{equation} \label{eq:2}
	\vspace{-2pt} 
	P_{u,t}  + P_{b,t} - P_{g,t} + P_{c,t} = 0
	\vspace{-2pt} 
\end{equation}

Assuming there is no loss in energy transmission and storage, and the charge and discharge efficiency $\eta $ is constant, we have:
\begin{equation}\label{eq:3}
	\vspace{-2pt} 
	E_{b,{t+1}}=\begin{cases}
		\begin{aligned}
			&E_{b,t} + \eta P_{b,t} \triangle t , &if \ P_{b,t}>0;\\ 
			&E_{b,t}, &if \  P_{b,t}=0;\\
			&E_{b,t} + \dfrac{P_{b,t} \triangle t}{\eta}, &if \  P_{b,t}<0.
		\end{aligned}
	\end{cases}
	\vspace{-2pt} 
\end{equation}
where $\triangle t$ is the time interval of a single charge or discharge action and $ E_{b,t}$ (kWh) is the state of charge (SOC) of battery. Moreover, the following should always be met in the above state transition:
\begin{equation}\label{eq:4}
	\vspace{-2pt} 
	{\small 
		\begin{cases}
			\begin{aligned}	
				&0 < P_{b,t}  \triangle t \leq \dfrac{E_b^{max}-E_{b,t}}{\eta},&if\ \ P_{b,t}>0;\\
				&(E_b^{min}-E_{b,t}) \eta \leq  P_{b,t} \triangle t < 0,&if \  P_{b,t}<0.
			\end{aligned}
	\end{cases}}
	\vspace{-1pt} 
\end{equation}
where $ E_b^{min} $ and $ E_b^{max} $ (kWh) are the minimum and maximum SOC, respectively.
\subsection{MDP Formulation}\label{sec2.2}
In this paper, an agent is used within a DRL framework to make appropriate control decisions on battery operations (charge, discharge or idle), while the environment is defined by the microgrid itself. The interaction between an agent and the microgrid can be described as a Markov Decision Process (MDP), which can be characterized by $(S, A, R, PM, \gamma) $ \cite{2b-1}. Here, $ S $ and $ A $ are the state and action spaces, $ R $ represents the accumulative reward from an action (the sum of immediate reward $ r_t $ and future reward $ R_{t+1} $); $ PM(s_{t+1} |s_t  ,a_t) $ indicates the state transition probability of the agent (moving to state $ s_{t+1} $ after taking action $ a_t\in A $ at state $ s_t\in S $); $ \gamma $ is the discount factor that aims to balance the rewards between $ r_t $ and $ R_{t+1} $ resulting from the action $ a_t $. A higher value of $\gamma$ means that the future reward is more valuable. Therefore, a microgrid MDP characterized by discrete actions and continuous states is defined as follows:

1) \textit{State space}: A continuous state space is defined by integrating the current status of controllable variable (state of charge $ E_{b,t} $) and the near-future status of uncontrollable variables ($ P_{u,t} $, $PR_t$ and $CI_t$):
\begin{equation} \label{eq:5}
		\vspace{-3pt} 
		s_t = (P_{u,t},\dots,P_{u,t+T};  PR_t,  \dots, PR_{t+T}; CI_t, \dots, CI_{t+T}; E_{b,t}) \raisetag{2.5\baselineskip}
	\vspace{-2pt} 
\end{equation}
where $ T $ represents the future time horizon that is taken into account for uncontrollable variables, and  $PR_t$ and $CI_t$ are the electricity price and carbon intensity, respectively. In this way, the near-future variations of the system can be factored to make the current control decision. This will involve the short-term prediction of the three uncontrollable variables, which is inevitably accompanied with uncertainties.

2) \textit{Action space}: A multi-level discrete action space is defined here to manage battery operations. In detail, the action $ a_t $ can be chosen from one of the following: full discharge, half discharge, idle, half charge, or full charge, represented by: {-1, -0.5, 0, 0.5, 1}. Then, the $ P_{b,t} $ can be computed as:
\begin{equation} \label{eq:6}
	\vspace{-5pt} 
	P_{b,t}=a_t E_{max}/\triangle t
\end{equation}
where $E_{max}$ (kWh) represents the maximum amount of energy that can be discharged/charged within $ \triangle t $.

3) \textit{Reward function}: Given the goal of reducing the operational costs and carbon emissions, the immediate reward $ r_t $ resulting from an action $a_t$ is defined as: 
\begin{align} 
	r_t &= [PR_t \max(P_{u,t},0) + \alpha CI_t \max(P_{u,t},0)]\triangle t \notag
	 \\ &- [PR_t \max(P_{u,t}+P_{b,t},0) + \alpha CI_t \max(P_{u,t}+P_{b,t},0) ]\triangle t \notag\\
	&=PR_t[\max(P_{u,t},0)-\max(P_{u,t}+P_{b,t},0)] \triangle t 
	\notag \\ & +  \alpha CI_t[\max(P_{u,t},0)-\max(P_{u,t}+P_{b,t},0)]\triangle t \label{eq:7}
\end{align}
where $ \alpha $ is a factor used to balance the importance between operational costs and carbon emissions. Here, $ [PR_t max(P_{u,t},0) + \alpha CI_t max(P_{u,t},0)]\triangle t $ represents the scenario where the battery is always inactive ($ P_{b,t} \equiv 0 $), while $  [PR_t max(P_{u,t}+P_{b,t},0)  + \alpha CI_t max(P_{u,t}+P_{b,t},0) ]\triangle t $ represents the scenario where the battery operates normally. Therefore, $ r_t $ can be considered as the aggregated reductions in operational costs and carbon emissions resulting from adopting the action $ a_t $ \cite{2b-2}. If $ r_t>0 $, it means that taking the action $ a_t $ yields positive results. Conversely, if $ r_t<0 $, it brings negative results. The objective of this paper is to use a DRL-based predictive control approach to find the best control strategy maximizing the reward.
\section{Predictive Control Approach Based on DRL}\label{sec3}
\begin{figure}[htbp]
	\vspace{-16pt} 
	\centerline{\includegraphics[width=9cm]{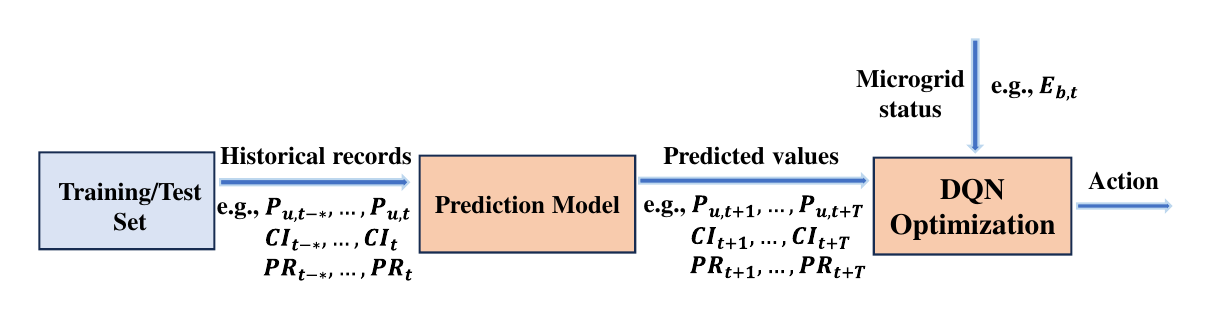}}
	\vspace{-12pt} 
	\caption{Framework of the DRL-based predictive control approach}
	\label{fig2}
	\vspace{-12pt} 
\end{figure}

Current DRL-based predictive control approaches overlook the issue of uncertainty caused by imperfect prediction models. This leads to suboptimal operational performance, particularly in the case of multi-step ahead predictions. To address this issue, this paper develops a new error temporal difference (ETD) algorithm for DRL to handle the prediction uncertainty. In the following sections, we will take the Deep Q Network (DQN)-based predictive control as an example to present the ETD algorithm and verify its effectiveness for DRL in optimizing microgrid operations. Given the generality of the TD algorithm in DRL, the new ETD algorithm can  also be applied to many other DRL methods. Fig. \ref{fig2} illustrates the framework of the DQN-based predictive control approach, which integrates the forecasts from a prediction model with the DQN. Here, the prediction model is first employed to infer the future values (e.g., $ P_{u,t+1},\dots,P_{u,t+T}$) of microgrid variables (i.e., $ P_u $, $ PR $ and $ CI $) based on the historical records. These forecasts are then combined with the battery status ($ E_{b,t} $) to be served as the state of the DQN for controlling battery operations. This paper focuses only on the DQN design that accounts for the prediction uncertainty, while the development of the prediction model is outside the scope of this work. 
\subsection{Traditional Temporal Difference Algorithm for DRL}\label{sec3.1}
Deep Q Network (DQN) \cite{2b-3} was developed to address the shortcomings of traditional reinforcement learning (RL) in handling complex, high-dimensional environments. It extends Q-learning \cite{2b-4}, one of the most popular RL methods, by leveraging deep neural networks (DNNs) to build complex mappings from high-dimensional inputs to action outputs. This enables DQN to tackle optimization problems with continuous state spaces that are beyond the capabilities of traditional RL \cite{2b-5}. Differing from Q-learning (which employs a lookup table to dynamically update and store the optimal reward value $  Q^*(s,a) $ of each action-state pair), DQN uses a DNN to approximate an optimal value function, i.e., $ Q(s_t,a_t;w)  = Q^*(s,a) $ where $ w $ is the network weights. Mathematically, the value function $ Q(s,a) $ is defined by the expectation of the cumulative discounted long-term reward as:
\begin{equation} \label{eq:8}
	\vspace{-2pt} 
	Q(s,a)= \mathbb{E}_\pi[R_t|s_t=s,a_t=a]
	\vspace{-2pt} 
\end{equation}
\begin{equation} \label{eq:9}
	\begin{split}
		R_t = \sum_{k=0}^{\infty} \gamma ^k  r_{t+k} = r_t + \gamma  R_{t+1}, \ \ \  \gamma \in [0,1]
	\end{split}
	\vspace{-2pt} 
\end{equation}
where $ k $ is the time index, and $ \mathbb{E} $ represents the Expectation function. In DQN, $ Q(s_t,a_t; w) $ can be learned through a series of interactions between the agent and the environment. At each time step (interaction), it is updated using the temporal difference (TD) algorithm \cite{2b-6}:
\begin{equation} \label{eq:10}
	\vspace{-5pt} 
		Q(s_t,a_t;w)\leftarrow Q(s_t,a_t;w) +l[r_t +\gamma max\underset{a}Q(s_{t+1},a;w^-)-Q(s_t,a_t;w)]
	\vspace{-3pt} 
\end{equation}
where $ l $ is the learning rate, $ w^- $ is the weights of the target network and $  r_t +\gamma max\underset{a}Q(s_{t+1},a;w^-) $ is called the TD target. 
\subsection{A New Error Temporal Difference Algorithm for DRL}\label{sec3.2}
A discount function $ \gamma $ is defined in DQN to balance the impact between future reward $ R_{t+1} $ and immediate reward $ r_t $ from an action. However, it does not specifically account for the uncertainty in the predicted values used in the context of predictive control. To address this, an additional discount factor for prediction error can be considered when employing predicted future values to update the Q network. Therefore, an error temporal difference (ETD) algorithm is proposed to update Q Network, as depicted in (\ref{eq:12}) below. Unlike the traditional TD, a new error discount factor $ \gamma_k' $ (at $k\text{-}$th time step) is introduced to weaken the effect of prediction uncertainty. Then, the Q Network of DQN can be updated via (\ref{eq:13}) at each time step.
\begin{equation} \label{eq:12}
	\vspace{-4pt} 
		R_t = \sum_{k=0}^{\infty} \gamma_k'  \gamma ^k  r_{t+k} = r_t + \gamma_k'  \gamma  R_{t+1}; \ \ \ \   \gamma, \gamma_k'\in [0,1]
	\vspace{-3pt} 
\end{equation}
\begin{equation} \label{eq:13}
	\vspace{-3pt} 
		Q(s_t,a_t; w) \leftarrow  Q(s_t,a_t; w) +  l[r_t +  \gamma_1' \gamma max\underset{a}Q(s_{t+1},a;w^-)-Q(s_t,a_t;w)]
	\vspace{-3pt} 
\end{equation}
Here, $ \gamma_k' $ is further defined as:
\begin{equation} \label{eq:14}
	\vspace{-2pt} 
	\gamma_k' = \begin{cases}
		\begin{aligned}
			& 1, & if \ k \leq 0;\\
			&(1-P_e^k)  \gamma_p, & if \ k \geq 1.
		\end{aligned}
	\end{cases}
	\vspace{-2pt} 
\end{equation}
where $ P_e^k $ is the combined uncertainty from all predicted microgrid variables at the $k\text{-}$th time step and $ \gamma_p $ is a constant parameter used to adjust $ P_e^k $. Here, $ P_e^k \in [0,1]$ can be obtained through preliminary experiments comparing the output of prediction models with actual records (to be presented later). $ 1-P_e^k $ can thus be regarded as the confidence of the predictions. For example, if $ P_e^k $ is 10\% at $ t +1 $ step, then we can get the confidence in the predictions as 90\%, thereby $ \gamma_1' = 0.9 \gamma_p$. Specifically, when $  k \leq 0 $, there is no prediction error as we can obtain the current sensor measurements, i.e., $ \gamma_k'  \equiv 1 $.  It can be observed that the determination of $ \gamma_p $ is crucial to compute $ \gamma_k' $. Before selecting the appropriate value of $ \gamma_p $, the related constraints are discussed first. Given the fact that the presence of prediction error must play a negative role in the learning process, thereby it should meet:
\begin{equation} \label{eq:15}
	\vspace{-2pt} 
	0 \leq \gamma_k' = (1-P_e^k)  \gamma_p \leq 1, \ \ \ \  k\geq 1
	\vspace{-2pt} 
\end{equation}
then:
\begin{equation} \label{eq:16}
	\vspace{-2pt} 
0 \leq \gamma_p \leq \frac{1}{1-P_e^k}, \ \ \ \  k\geq 1
	\vspace{-2pt} 
\end{equation}

Furthermore, when conducting multi-step predictions, the prediction error can increase significantly as the number of prediction steps grows (to be discussed in Section \ref{sec4.1}). Applying the confidence ($ 1-P_e^k $) directly as the error discount factor $ \gamma_k' $ may lead to over-discounting. The design of $\gamma_p$ is to ensure that the error discount is within a reasonable range. Given this, $\gamma_p$ should be defined as a number greater than 1, which gives the final constraint as:
\begin{equation} \label{eq:17}
	\vspace{-2pt} 
	1 \leq \gamma_p \leq \frac{1}{1-P_e^k}, \ \ \  k\geq 1
	\vspace{-2pt} 
\end{equation}

In this paper, we choose $\gamma_p$  as the center of the bound constraint $[1, 1/(1-P_e^k )]$, namely:
\begin{equation} \label{eq:18}
	\vspace{-3pt} 
	\gamma_p = \frac{1}{2}(1 + \frac{1}{1-P_e^k}) = \frac{2-P_e^k}{2(1-P_e^k)}, \ \ \ \  k\geq 1
	\vspace{-1pt} 
\end{equation}

Then, substituting (\ref{eq:18}) into (\ref{eq:14}), this will give:
\begin{equation}\label{eq:19}
	\vspace{-3pt} 
	\gamma_k' = \begin{cases}
		\begin{aligned}
			& 1, &if \  k\leq 0;\\
			&\frac{2-P_e^k}{2}, &if\   k\geq 1.\\			
		\end{aligned}
	\end{cases}
\end{equation}

In this way, $ \gamma_k' $ can be computed adaptively once $ P_e^k $ is known. The lower the prediction error, the more reliable the predicted value becomes, resulting in a larger $ \gamma_k' $. In particular, when the prediction accuracy is 100\% ($ P_e^k=0 $), then  $ \gamma_k' =1 $ will hold, which means no error discount and the ETD degenerates into the traditional TD algorithm. Now, how to get the value of $ P_e^k $ becomes a key issue. However, it is not always a straightforward task, especially when multiple variables (e.g., $ P_{u,t}, PR_t $ and $ CI_t $ are predicted. Generally, different variables will also have different prediction errors. Moreover, their contributions to the reward function also vary during the Q network optimization process, as depicted in (\ref{eq:7}). Given this, the weighted average method is employed in this paper to quantify the value of $ P_e^k $.

It can be seen from (\ref{eq:7}) that $ PR_t $ only affects operational costs and $ CI_t $ only affects carbon emissions, while $ P_{u,t} $ affects both of these two items. Additionally, the contribution of operational costs and carbon emissions to the reward can be represented as $ 1: \alpha$. Given these, the following can be derived:
\begin{equation} \label{eq:20}
	\vspace{-2.5pt} 
		C_u : C_{c} : C_{p}= 1+\alpha : \alpha : 1 = \dfrac{1}{2}:\dfrac{\alpha}{2+2\alpha}:\dfrac{1}{2+2\alpha} 
	\vspace{0pt} 
\end{equation}
where $ C_u, C_c $, and $  C_p $ are the contribution factor of the $ P_{u,t}, PR_t $ and $ CI_t $ to the reward. Then, $ P_e^k $ can be quantified as:
\begin{equation} \label{eq:21}
	\vspace{-2.5pt} 
	P_e^k = C_u  MPE_u + C_{c}  MPE_{c} +  C_{p}MPE_{p}, \ \ \  k\geq 1
	\vspace{-1pt} 
\end{equation}
where $MPE_u, MPE_{c}$ and $MPE_{p}$  are the mean percentage errors (MPE) of the predictions for $ P_{u,t}, CI_t$ and $ PR_t $, respectively. Once $ P_e^k $ is obtained, $ \gamma_k' $ can be computed using (\ref{eq:19}). 
%

The complete steps of the DQN-based predictive control approach are depicted below. Historical measurement records are first sent to the prediction model to forecast the near-future values of the three variables ($ PR $, $ C $I and $ P_u $). These predicted values, along with their MPE, are then fed into DQN to find the optimal control strategy. There are a total of $ N_m $ episodes in the DQN optimization. In each episode, a series of $ K $ time series sequences are sampled randomly, with each time series sequence containing $ 3 (T+1 $) values. These values are later combined with the battery status ($ E_{b,t} $) to form the state vector of DQN. Consequently, the agent engages in a series of MDP interactions ($ K $ times) with the microgrid environment. During each interaction, the action is chosen based on the optimized policy and $ \epsilon $-greedy \cite{2b-6-1} and  $ Q(s_t,a_t; w) $ is updated using the new ETD algorithm (\ref{eq:13}). After completing $ N_m $ episodes, the optimal policy will be obtained. 

\section{Simulations}\label{sec4}
\vspace{-5pt} 
\subsection{Performance Evaluation}\label{sec4.1}

The developed ETD was evaluated using an open dataset containing one year of hourly loads, PV generation and electricity price for a district microgrid system in the US \cite{2b-6-2,2b-6-2-1}. The corresponding carbon intensity data was downloaded from the WattTime website \cite{2b-6-3}. The complete dataset can be accessed via \cite{2b-2}. Here, we added Gaussian noise with a variance of 5\% to the original unmet power to simulate the microgrid noises in sensor measurements. Two unmet power datasets are thus generated independently: one is used for model training while the other is for testing. In the following sections, the actual values always refer to data with noises, while original unmet power data is only used here as a baseline. We further assume that the carbon intensity and electricity price collected are noise free. In this paper, we trained DQN using week-long (168 hours, i.e., $  K=168 $) time series sequences per episode and the next 6-hour predictions ($ T=6 $) for each variable at each time step. The minimum and maximum SOC were defined as: $ E_b^{min} = 0.2E_b^c $ and $ E_b^{max} = E_b^c $, where $ E_b^c =1000 $(kWh) represents the capacity of the battery. The balance factor $ \alpha $ in (\ref{eq:7}) was defined as 0.25. Moreover, a DQN with three fully connected layers, each containing 64 neurons, was used for optimization. The optimizer employed was Adam, the activation function was ReLU and the batch size was 64. Through trials and errors, other parameter settings were set as: $ \gamma=0.99, l=0.0001, N_m = 5000$ and $\epsilon = 0.1 $. All simulations were conducted using Python and the Keras library on Google Colab without any hardware accelerator.

The evaluation was conducted based on the outputs from two different prediction models: CNN-LSTM and SOIT2FNN-MO. For detailed information about these models, please refer to our previous works \cite{1-12} and \cite{1-13}. Each model makes predictions for the next six time steps based on the measurements from previous 12 time steps, the prediction errors (MPE) being presented in Table \ref{tab1} and Table \ref{tab2}. It can be seen that considerable prediction errors can occur, especially when the predicted steps increase. 
\begin{table}[htbp]
	\vspace{-15pt} 
	\caption{Six-step ahead prediction errors (MPE) using CNN-LSTM \cite{1-12}}
	\vspace{-2pt} 
	\centering
	\renewcommand{\arraystretch}{1.1} 
	\setlength{\tabcolsep}{3pt} 
	\begin{tabular}{|c|c|c|c|c|c|c|}
		\hline
		\multirow{2}{*}{Item} & \multicolumn{6}{c|}{Prediction steps} \\
		\cline{2-7}
		& 1 & 2 & 3 & 4 & 5 & 6 \\
		\hline
		$ CI $  & 6.684 &14.25 &20.19 &33.87 &34.71 &38.52\\
		\hline
		$ PR $  &  9.982 &13.56 &19.71 &26.00 &33.90 &32.17 \\
		\hline
		$ P_{u,t} $  &  14.26 &17.20 &20.55 &26.38 &32.15 &34.69 \\
		\hline
	\end{tabular}
	\label{tab1}
	\vspace{-2pt} 
\end{table}
\begin{table}[htbp]
	\vspace{-2pt} 
	\caption{Six-step ahead prediction errors (MPE) using SOIT2FNN-MO
		\cite{1-13}}
	\vspace{-2pt} 
	\centering
	\renewcommand{\arraystretch}{1.1} 
	\setlength{\tabcolsep}{3pt} 
	\begin{tabular}{|c|c|c|c|c|c|c|}
		\hline
		\multirow{2}{*}{Item} &  \multicolumn{6}{c|}{Prediction steps} \\
		\cline{2-7}
		& 1 & 2 & 3 & 4 & 5 & 6 \\
		\hline
		$ CI $ & 7.847 &17.38 &27.32 &35.28 &27.32 &37.42\\
		\hline
		$ PR $ &  11.03 &18.52 &24.14 &28.42 &31.96 &34.82 \\
		\hline
		$ P_{u,t} $ &  12.19 &19.12 &24.75 &27.88 &29.40 &29.76 \\
		\hline
	\end{tabular}
	\label{tab2}
	\vspace{-12pt} 
\end{table}

\vspace{-14pt} 
After obtaining predicted values and MPEs, the ETD algorithm was evaluated based on the above predictions. Here, the DQN model was trained in two cases: 1) actual data was simply used as the predictions; 2) outputs from the above model was used as predictions. On the other hand, the evaluation of the trained DQNs was conducted solely using predicted values from the model output, as in real-world scenarios they can only be obtained through predictions. The comparisons of annual cumulative rewards (ACR) obtained using different trained models on the test set are listed in Table \ref{tab3}, where \textit{Pred 1} and \textit{Pred 2} represent the values predicted by CNN-LSTM and SOIT2FNN-MO, respectively. It can be observed that the developed ETD-DQN approach performed better than the traditional DQN on both two prediction models, confirming the effectiveness of the ETD algorithm. Additionally, an interesting phenomenon is that DQN training with predicted values yielded better results than that with the actual data. This is reasonable because incorporating predicted values into training can account for prediction errors during learning, thereby increasing the control resilience to the uncertainty.
\begin{table}[htbp]
	\vspace{-10pt} 
	\caption{Annual cumulative rewards (ACR) obtained using different trained models}
	\vspace{-12pt} 
	\begin{center}
		\renewcommand{\arraystretch}{1.1} 
		\setlength{\tabcolsep}{3pt} 
		\begin{tabular}{|c|c|c|c|c|c|c|c|c|}
			\hline
			\textbf{Model} & \multicolumn{4}{|c|}{\textbf{DQN}} & \multicolumn{4}{|c|}{\textbf{ETD-DQN}}\\
			\hline
			Train & \multicolumn{2}{|c|}{\textbf{Actual}} & Pred 1 & Pred 2 & \multicolumn{2}{|c|}{\textbf{Actual}} & Pred 1 & Pred 2\\
			\hline
			Test & Pred 1 & Pred 2 & Pred 1 & Pred 2 & Pred 1 & Pred 2 & Pred 1 & Pred 2\\
			\hline
			ACR & 161.4 & 157.6 & 165.7 & 163.7 & 165.3 & 164.0 & 170.1 & 174.7\\
			\hline
		\end{tabular}
		\label{tab3}
		\vspace{-40pt} 
	\end{center}
\end{table}
\subsection{Objective Evaluation and Visualization}\label{sec4.2}
\begin{figure}[htbp]
	\vspace{-10pt} 
	\centerline{\includegraphics[width=12cm]{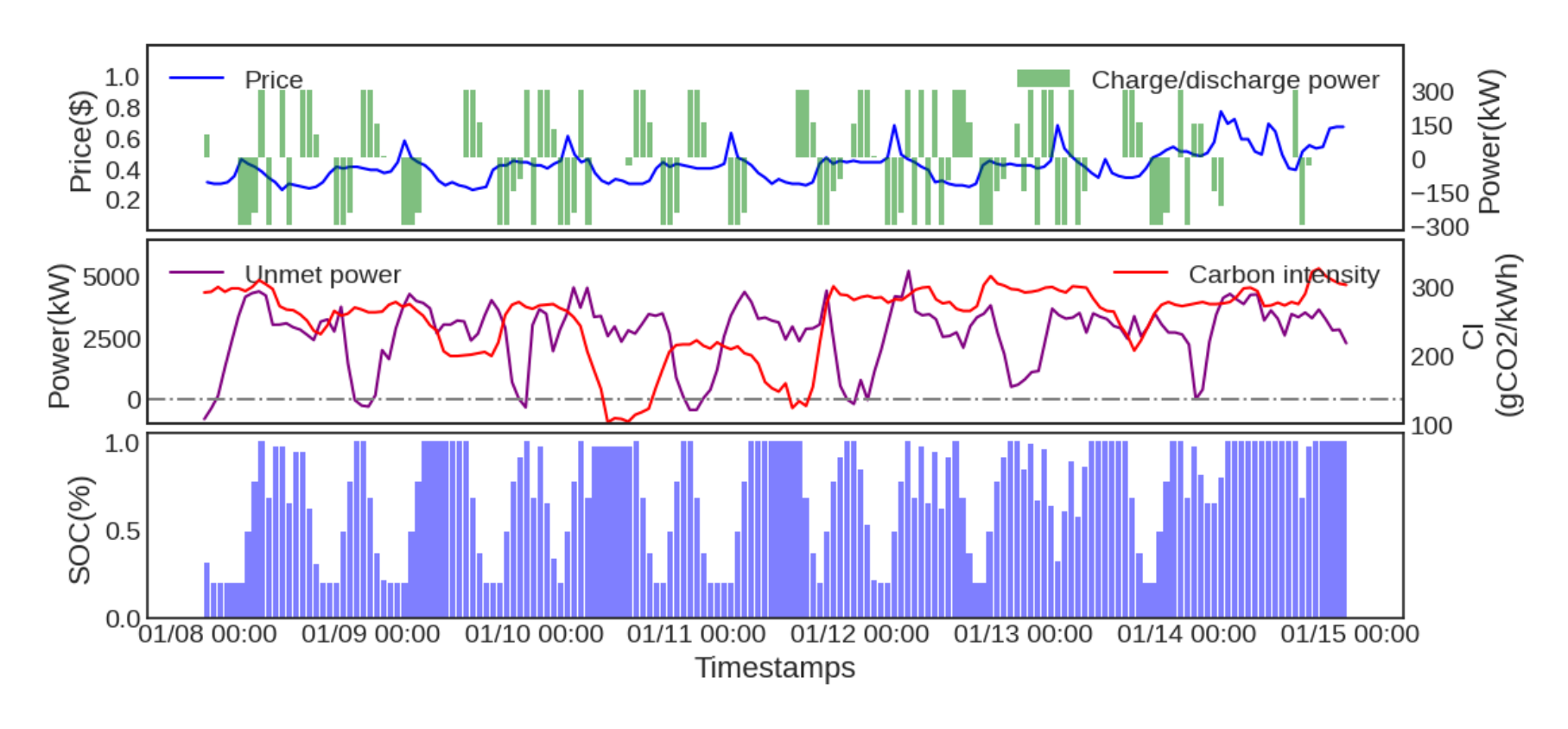}}
	\vspace{-13pt} 
	\caption{A week-long battery control simulation}
	\label{fig4}
	\vspace{-20pt} 
\end{figure}

This section also evaluates the actual reductions in operational costs and carbon emissions using DQN and ETD-DQN. The results demonstrated that ETD-DQN outperforms DQN in reducing both operational costs and carbon emissions. When using CNN-LSTM as the prediction model, the ETD-DQN approach reduced annual operational costs by \$136.0k and carbon emissions by 49.5 tCO2. This represents a 3.19\% greater reduction in costs and a 1.43\% greater reduction in emissions compared to the DQN, which achieved annual reductions of \$131.8k in costs and 48.8 tCO2 in emissions. Similarly, with SOIT2FNN-MO as the prediction model, ETD-DQN decreased annual operational costs by \$139.2k and carbon emissions by 51.1 tCO2. This corresponds to a 7.41\% higher reduction in costs and a 3.87\% higher reduction in emissions than the DQN (with reductions of \$129.6k in costs and 49.2 tCO2 in emissions).Given this, we further visualized a week of battery control simulations using ETD-DQN based on predictions from the SOIT2FNN-MO, as shown in Fig. \ref{fig4}. The visualization shows that the microgrid effectively made the expected charging/discharging decisions. Specifically, when PV generation exceeded demand, the microgrid stored the excess power in the battery for future use. Finally, when PV generation fell short of demand: i) if carbon intensity and electricity price were low, the microgrid imported a significant amount of power from the grid to charge the battery; ii) if carbon intensity and the price were high, the microgrid prioritized using the stored battery power to minimize unmet power demand.
\section{Conclusion}\label{sec5}
This paper presented a new ETD algorithm for DRL in optimizing microgrid operations, aiming to address the uncertainty caused by imperfect prediction models. A weighted average was devised to quantify the prediction uncertainty based on the known prediction errors. Then, the ETD algorithm was proposed to effectively incorporate the uncertainty information in the DQN learning process. Finally, a series of simulations were conducted on a typical microgrid system integrated with RES and ESS. Experimental results demonstrated that the ETD effectively enhanced the performance of DQN in optimizing microgrid operations. Considering the wide applicability of the TD algorithm in DRL, the new ETD algorithm will be generalized to other DRL methods in the future.

\end{document}